\documentclass[letterpaper, 10 pt, conference]{ieeeconf} 

\IEEEoverridecommandlockouts                        % This command is only needed if 
\usepackage{graphics}
\usepackage{amsmath} 
\usepackage{amssymb}
\usepackage{subcaption}
\usepackage{wrapfig}
\usepackage{cite}
\usepackage{float}
\usepackage{svg}
\usepackage{tikz}
\usepackage{tabularx}
\usepackage{array}
\usepackage[font=small]{caption}
\usepackage{booktabs}
\usepackage{bm}
\usepackage{algorithm}
\usepackage{algorithmic}
\usepackage{bbm}
\usepackage{verbatim}
\usepackage{multirow}
\usepackage{pgfplots}
\let\labelindent\relax % https://tex.stackexchange.com/questions/170772/command-labelindent-already-defined
\usepackage[inline]{enumitem}
\PassOptionsToPackage{hyphens}{url}
\usepackage[colorlinks=true, citecolor=magenta]{hyperref}

\newtheorem{definition}{Definition}

\usepackage{tablefootnote}

\DeclareMathOperator*{\argmax}{argmax}
\DeclareMathOperator*{\argmin}{argmin}

\definecolor{green}{RGB}{11,155,13}

\title{
Rethinking Social Robot Navigation: Leveraging the Best of Two Worlds
}

\author{Amir Hossain Raj$^{1*}$, Zichao Hu$^{2*}$, Haresh Karnan$^2$, Rohan Chandra$^2$, Amirreza Payandeh$^1$, Luisa Mao$^2$, \\Peter Stone$^{2,3}$, Joydeep Biswas$^2$, and Xuesu Xiao$^1$
\thanks{$^*$Equally contributing authors. $^1$ George Mason University. $^2$The University of Texas at Austin. $^3$Sony AI. {\tt\scriptsize \{araj20, apayande, xiao\}@gmu.edu, \{zichao, haresh.miriyala, rchandra, luisa.mao\}@utexas.edu, \{pstone, joydeepb\}@cs.texas.edu} 
}
}

\begin{document}
\makeatletter
\g@addto@macro\@maketitle{
  \begin{figure}[H]
  \setlength{\linewidth}{\textwidth}
  \setlength{\hsize}{\textwidth}
  \centering
  \includegraphics[width=\textwidth]{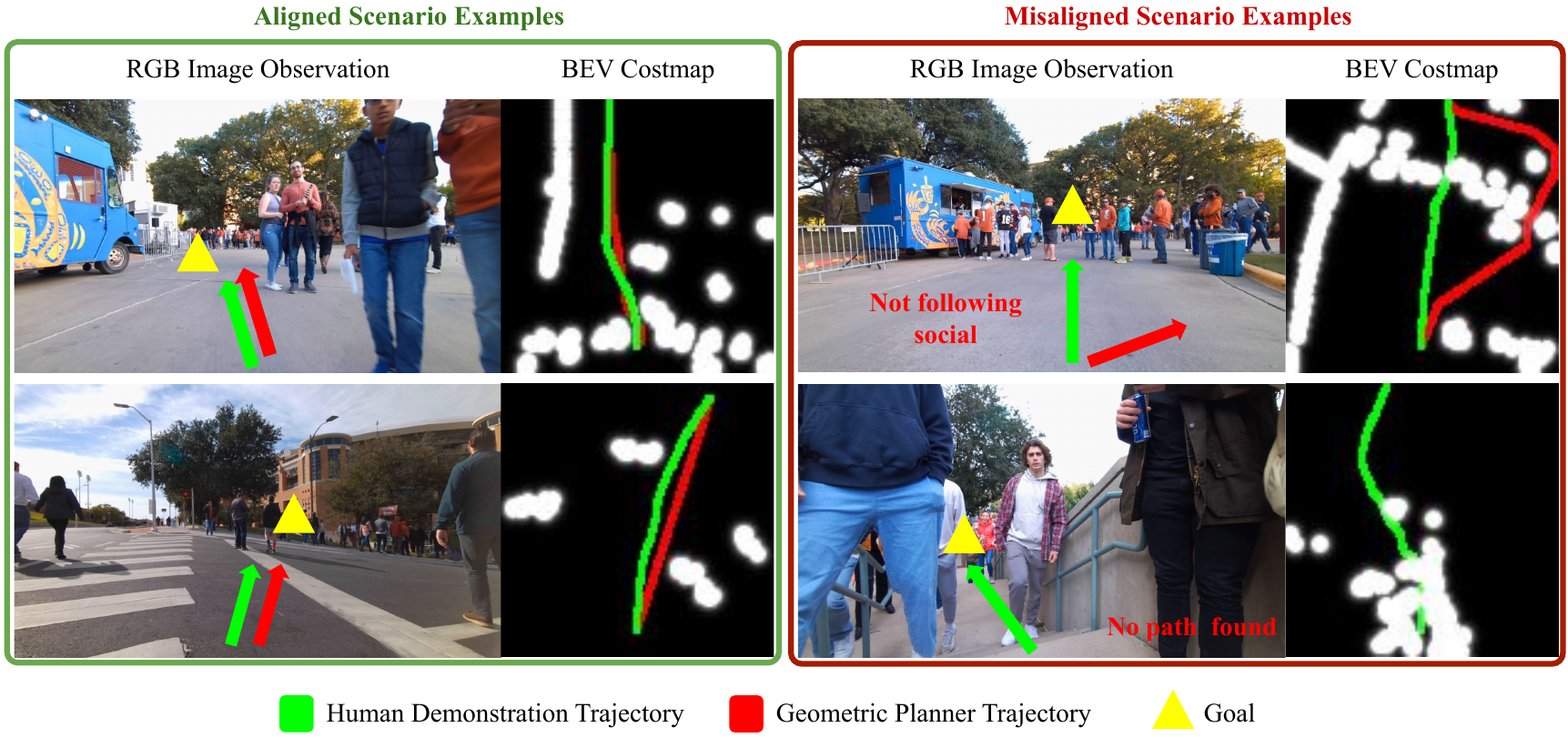}
  \caption{Example Scenarios for Geometric Plans vs. Human Demonstrations in the \textsc{scand} dataset: In many social scenarios, the geometric planner can produce trajectories that align with human demonstration trajectories (left). While uncommon, it is inevitable for geometric planners to encounter social scenarios that they cannot handle (right). For example, the robot should approach the waiting line carefully and expect people to make room for it to pass (top right) and carefully move forward after the person on the left passes (bottom right).
    }
    \label{fig:cover}
  \end{figure}
}
\makeatother
\maketitle
\addtocounter{figure}{-1}
% \thispagestyle{fancy}
% \pagestyle{empty}

%%%%%%%%%%%%%%%%%%%%%%%%%%%%%%%%%%%%%%%%%%%%%%%%%%%%%%%%%%%%%%%%%%%%%%%%%%%%%%%%
\begin{abstract}

Empowering robots to navigate in a socially compliant manner is essential for the acceptance of robots moving in human-inhabited environments. Previously, roboticists have developed geometric navigation systems with decades of empirical validation to achieve safety and efficiency. However, the many complex factors of social compliance make geometric navigation systems hard to adapt to social situations, where no amount of tuning enables them to be both safe (people are too unpredictable) and efficient (the frozen robot problem). With recent advances in deep learning approaches, the common reaction has been to entirely discard these classical navigation systems and start from scratch, building a completely new learning-based social navigation planner. In this work, we find that this reaction is unnecessarily extreme: using a large-scale real-world social navigation dataset, \textsc{scand}, we find that \textit{geometric systems can produce trajectory plans that align with the human demonstrations in a large number of social situations}. We, therefore, ask if we can rethink the social robot navigation problem by leveraging the advantages of both geometric and learning-based methods. We validate this hybrid paradigm through a proof-of-concept experiment, in which we develop a hybrid planner that switches between geometric and learning-based planning. Our experiments on both \textsc{scand} and two physical robots show that the hybrid planner can achieve better social compliance compared to using either the geometric or learning-based approach alone.

\end{abstract}

%%%%%%%%%%%%%%%%%%%%%%%%%%%%%%%%%%%%%%%%%%%%%%%%%%%%%%%%%%%%%%%%%%%%%%%%%%%%%%%%
\section{Introduction}
\label{sec::introduction}
% need to move this to first page
% \begin{figure*}[t]
%     \centering
%     \includegraphics[width=\textwidth]{figure/scand_fig1.pdf}
%     \caption{Example Scenarios for Geometric Plans vs. Human Demonstrations in the \textsc{scand} dataset: In many social scenarios, the geometric planner can produce trajectories that align with human demonstration trajectories (left). While uncommon, it is inevitable for geometric planners to encounter social scenarios that they cannot handle (right). For example, the robot should approach the waiting line carefully and expect people to make room for it to pass (top right) and carefully move forward after the person on the left passes (bottom right).
%     }
%     \label{fig:cover}
% \end{figure*}

Decades of research into autonomous mobile robot navigation allows robots to reliably move from one point to another without collision with (mostly static) obstacles~\cite{fox1997dynamic, quinlan1993elastic, xiao2022autonomous, perille2020benchmarking, nair2022dynabarn}.
Recently, there is a growing interest in bringing robots out of academic labs and into common public spaces in the wild~\cite{scout, starship, dilligent, tinymile}. On their way to deliver packages~\cite{scout}, take-outs~\cite{starship}, and medical supplies~\cite{dilligent}, those robots need to navigate in a way such that they not only avoid static obstacles and move towards their goals, but also take other pedestrian's objective into account. 
Therefore, enabling robots to navigate by social norms, known as the social robot navigation problem, has emerged as an important research topic.

To address social robot navigation, researchers have collected large-scale demonstration datasets~\cite{karnan2022socially, scand, nguyen2023toward, francis2023socialhri, francis2023principles}, created protocols to validate social navigation systems~\cite{pirk2022protocol, francis2023socialhri, francis2023principles}, and developed social navigation techniques using geometric~\cite{kirby2010social, mirsky2021conflict, hart2020using, singamaneni2021human, kollmitz15ecmr, chandra2023decentralized, chandra2023socialmapf, bilevel, holtz2021iterative} or learning-based~\cite{rosmann2017online, chen2017socially, karnan2022voila, xiao2022motion, xiao2022learning, park2023learning, holtz2021idips, holtz2022socialgym, sprague2023socialgym} approaches to move robots in a safe and socially compliant manner. While geometric approaches enjoy safety, explainability, and certifiabiliy, they require extensive engineering effort and are not scalable to complex and diverse social scenarios. On the other hand, learning-based approaches conveniently enable social navigation behaviors in a data-driven manner, but forfeit most of the benefits of their geometric counterparts. Moreover, most of these approaches have achieved improvement in social compliance primarily in experiments conducted in controlled lab environments. 

Despite such academic successes, robotics practitioners are still reluctant in deploying those state-of-the-art social navigation systems on their robot fleet in the real world, especially data-driven approaches, due to their lack of safety, explainability, and testability. To the best of our knowledge, most of the navigation stacks running on real-world service robots are still geometric systems, which can be rigorously tested, confidently deployed, and easily debugged from a large-scale, real-world, software engineering perspective. 

% \begin{figure}[t]
%      \centering
%          \includegraphics[width=1\columnwidth]{figure/1-2F.png}
%          % \includegraphics[width=0.48\columnwidth]{figure/2F-LARGE-.png}
%         \caption{Comparison of the navigation behavior from \textsc{scand} demonstration, \texttt{move\_base}, and our approach at the same time step in two different social scenarios: Classical \texttt{move\_base} (red) aligns with (left) or deviates from (right) the \textsc{scand} demonstration, while our approach is always close to the socially compliant demonstration. \zichao{I am not a big fan of this image since the hybrid planner is not what we are trying to sell in this paper. Rather I propose we include a few images to show some social compliant scenarios vs some socially incompliant scenarios.}}
%         \label{fig:cover}
% \end{figure}

Considering the stark contrast between (i) our decade-long research and the current public resistance to mobile robots in public spaces and (ii) the active academic research in social robot navigation and the industrial reluctance to use them in real-world practices, we present a case study on social compliance of different existing robot navigation systems using a state-of-the-art Socially CompliAnt Navigation Dataset (\textsc{scand})~\cite{karnan2022socially, scand} as a benchmark. We discover from the case study that geometric navigation systems can produce trajectories that align with human demonstrations in many social situations (up to $80\%$, e.g., Fig.~\ref{fig:cover} left). 
This result motivates us to rethink the approach to social robot navigation by adopting a hybrid paradigm that leverages both geometric-based and learning-based methods. We validate this paradigm by designing a proof-of-concept experiment in which we develop a hybrid planner that switches between geometric and learning-based planning. 
Our main contributions are:
\begin{itemize}
    \item A social compliance definition based on how well a navigation behavior aligns with a human demonstration. 
    \item A case study on the \textsc{scand} dataset which provides evidence that 
    \begin{enumerate*}
    \item geometric navigation systems can produce socially compliant navigation behaviors in many social scenarios (up to 80\%, Fig.~\ref{fig:cover} left) but will inevitably fail in certain social scenarios (Fig.~\ref{fig:cover} right); and
    \item learning-based methods have the potential to solve challenging social scenarios where geometric approaches fail but using them alone can suffer from distribution-shift problems. 
    \end{enumerate*}
    \item A hybrid paradigm that leverages both geometric-based and learning-based methods for social robot navigation. We validate this paradigm using both playback tests on \textsc{scand} and physical robot experiments on two robots.
\end{itemize}

\section{Background}
\label{sec::related_work}
% In this section, we review various approaches to mobile robot navigation, focusing on both geometric and learning-based methods. Additionally, we examine the \textsc{scand} dataset as a part of our review.

\subsection{Classical Geometric Navigation}
\label{subsec: related_classical}
As a research topic since decades ago, roboticists have developed a plethora of classical navigation systems to move robots from one point to another without collision with obstacles. 
Most classical systems take a global path from a high-level global planner, such as Dijkstra's~\cite{dijkstra1959note}, A*~\cite{Hart1968}, or D*~\cite{ferguson2006using} algorithm, and seek help from a local planner~\cite{fox1997dynamic, quinlan1993elastic} to produce fine-grained motion commands to drive robots along the global plan and avoid obstacles. 
Most classical navigation systems require a pre-defined cost function~\cite{lu2014layered} for both global and local planning and trade off different aspects of the navigation problem, such as path length, obstacle clearance, energy consumption, and in recent years, social compliance. 
They then use sampling-based~\cite{kavraki1996probabilistic, kuffner2000rrt, fox1997dynamic}, optimization-based~\cite{ratliff2009chomp, quinlan1993elastic}, or potential-field-based~\cite{koren1991potential} methods to generate motion commands. 
These approaches enjoy benefits such as safety, explainability, and testability, which can be provably or asymptotically optimal. Such benefits are important when deploying physical robots in the real world with humans around, and therefore these classical navigation systems are still widely favored by practitioners in the robot industry~\cite{scout, starship, dilligent, tinymile}. 
Implementing classical navigation approaches in socially challenging environments, however, requires substantial engineering effort such as manually designing cost functions~\cite{lu2014layered} or fine-tuning navigation parameters~\cite{zheng2021ros}. These drawbacks motivate the use of learning-based approaches for the social robot navigation problem.

\subsection{Learning-Based Navigation}
\label{subsec: related_learning}
Learning-based approaches~\cite{xiao2022motion, baghaei2022deep} may be either end-to-end~\cite{pan2020imitation}, i.e., producing actions directly from perception, or in a structured fashion, e.g., learning local planners~\cite{faust2018prm, chiang2019learning, francis2020long, xiao2021toward, xiao2021agile, wang2021agile, liu2021lifelong, xu2023benchmarking}, cost functions~\cite{perez2014robot, wigness2018robot, xiao2022learning, sikand2022visual}, kinodynamics models~\cite{xiao2021learning, karnan2022vi, atreya2022high}, and planner parameters~\cite{teso2019predictive, bhardwaj2019differentiable, xiao2022appl, xiao2020appld, wang2021appli, wang2021apple, xu2021applr, xu2021machine}. From the learning perspective, most approaches fall under either imitation learning~\cite{pan2020imitation, wigness2018robot, xiao2020appld, karnan2022voila, xiao2022learning} from expert demonstrations or reinforcement learning~\cite{faust2018prm, chiang2019learning, francis2020long, xu2021applr, xu2023benchmarking, xu2021machine} from trial and error. Despite the convenience of learning emergent navigational behaviors purely from data in social scenarios, these systems suffer from the lack of safety and explainability, and cannot easily go through rigorous software testing and be debugged and fixed to avoid future failure cases. Therefore, robot practitioners rarely use learning-based navigation systems in their robot fleets deployed in the real world. 

\subsection{\textsc{scand}}
\textsc{scand}~\cite{karnan2022socially, scand} contains $8.7$ hours, $138$ trajectories, and $25$ miles of socially compliant, human tele-operated robot navigation demonstrations on the busy campus of The University of Texas at Austin, USA. \textsc{scand} includes socially challenging scenarios such as following, intersection, and overtaking, making it an ideal dataset to test social robot navigation methods. Additionally, \textsc{scand} provides multi-modal information, including 3D LiDAR, RGB images, joystick commands, odometry, and inertia readings, collected on two morphologically different mobile robots---a Boston Dynamics Spot and a Clearpath Jackal---controlled by four human demonstrators in indoor and outdoor environments.

% \begin{table*}[t]
%     \centering
%     \begin{tabular}{rcc}
%     \toprule
%        Navigation  & Benefits & Limitations\\
%        \midrule
%         Classical (Section~\ref{subsec: related_classical}) & Robustness, safety, completeness &Manual tuning required\\
%         Learning-based (Section~\ref{subsec: related_learning}) & No manual tuning required & Sim-to-real gap\\
%         \bottomrule
%     \end{tabular}
%     \caption{\textbf{Classical vs Learning-based Methods:} We summarize the benefits and limitations of classical and learning-based approaches for social navigation.}
%         % \vspace{-15pt}
%     \label{tab: classical_vs_learning}
% \end{table*}

\section{Social Compliance Case Study}
In our case study, we first propose a definition of social compliance based on how well a navigation behavior aligns with the human demonstration. Then, we use this metric to benchmark the social compliance of a set of geometric navigation systems on \textsc{scand}. Three major findings from these analyses are:
\begin{enumerate}
    \item Different geometric navigation systems show similar performance, where their trajectories align with human demonstrations in up to 80\% \textsc{scand} scenarios.
    \item A general purpose navigation planner, e.g., \texttt{move\_base}, even without tuning for specific scenarios can often produce trajectories aligning with human demonstrations.
    \item Despite the inevitable failures of geometric navigation planners that cannot be resolved through manual tuning, such occurrences remain infrequent within \textsc{scand}.
\end{enumerate}
Finally, we investigate the performance of geometric-based and learning-based planners on both an in-distribution and out-of-distribution test set and find that training on \textsc{scand} can suffer from distribution-shift problems.

\subsection{Defining Social Compliance on \textsc{scand}}
\textsc{scand} is a dataset designed for learning from demonstration research, so we assume the human demonstrations provided in the dataset to be the ground truth\footnote{Human demonstrations may not always represent the sole socially compliant behavior in certain scenarios, thereby challenging our assumption that deviations from the ground truth inherently lack social compliance~\cite{karnan2022socially}.}. In this work, we propose a definition of social compliance based on how well a navigation behavior produced by a navigation system aligns with the human demonstration. 

\begin{definition}
Given a navigation scenario $\mathcal{S}_t$ with a human demonstration behavior $\mathcal{B}^{D}_t$ at a time step $t$, a navigation behavior $\mathcal{B}_t$ is \textbf{socially compliant} if $d = \lVert \mathcal{B}_t-\mathcal{B}^{D}_t \rVert_{\textsc{d}} < \epsilon$, where $\lVert\cdot\rVert_\textsc{d}$ is a distance metric and $\epsilon$ is a small threshold.
\label{def: social_compliant}
\end{definition}

In this work, we define the navigation scenario $\mathcal{S}_t$ to be the robot sensor observations in the past two seconds at a certain time step $t$ of a demonstration trajectory in \textsc{scand}. We also define human demonstration behavior $\mathcal{B}^{D}_t$ as a deterministic sequence of waypoints, $P^D_i=(x^D_i,y^D_i)$, at every step $i$ starting from time step $t$, to a navigation goal, $P^D_{t_g}$, $10$m ahead of the robot at time step $t_g$ on the human demonstrated path: $\mathcal{B}^{D}_t = \{P^D_i\}_{i=t}^{t_g}$, where $t_g$ is the first time step when $\lVert P^D_{t} -P^D_{t_g})\rVert\geq 10$.
 % \begin{equation}
 %      \mathcal{B}^{D}_t = \{P^D_i | i\in [t, t_g], t_g = t+\argmin_{i}\lVert P^D_{t} -P^D_{t+i})\rVert\geq 10\}
 % \end{equation}
Then we send the same navigation goal, $P^D_{t_g}$, to a navigation system to be benchmarked and retrieve its planned trajectory $\mathcal{B}_t$. Finally, considering most geometric navigation systems use a global trajectory planner before local motion planning and the quality of the robot motions heavily depends on the quality of the planned global trajectories $\mathcal{B}_t$, we use the undirected Hausdorff distance, which measures the distance from each point in the planned trajectory $\mathcal{B}_t$ to the closest point in the corresponding demonstration trajectory $\mathcal{B}^D_t$ as $\lVert\cdot\rVert_\textsc{d}$, to measure the social compliance of the planned trajectory:
\begin{align}
\begin{split}
     d = \max\{
     &\sup_{P^D_i \in \mathcal{B}^{D}_t} \inf_{P_i \in \mathcal{B}_t} \lVert P^D_{i} - P_{i} \rVert, \\
     &\sup_{P_i \in \mathcal{B}_t} \inf_{P^D_i \in \mathcal{B}^{D}_t} \lVert P^D_{i} - P_{i} \rVert \}. \nonumber
\end{split}
\end{align}

\begin{figure}[t]
    \centering
    \includegraphics[width=8.5cm]{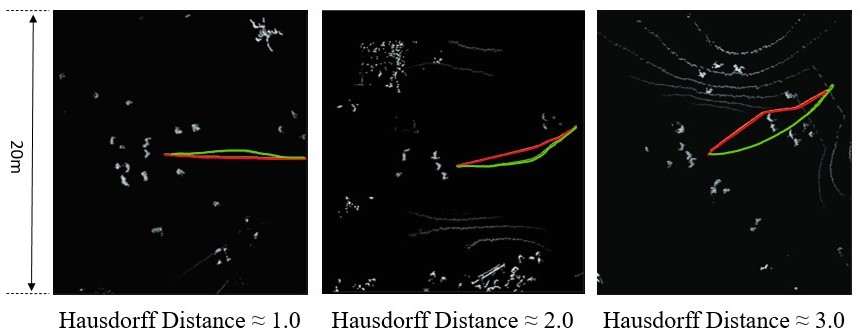}
    \caption{Different Hausdorff distances between the human demonstration trajectory (green) and geometric planner trajectory (red). White dots denote nearby humans and obstacles. Empirically, navigation systems that align better with human demonstrations are expected to yield a lower Hausdorff distance. }
    \label{fig:HdistanceComparison}
\end{figure}

% Considering most geometric navigation systems use a global trajectory planner before local motion planning and the quality of the robot motions heavily depends on the quality of the planned global trajectories $\mathcal{B}_t$, we use Hausdorff distance measures how similar two trajectory is a reasonable metric to benchmark the performance of each system. 
Fig.~\ref{fig:HdistanceComparison} shows trajectories with different Hausdorff distances from a Bird's Eye View (BEV). Empirically, navigation systems that align better with human demonstrations are expected to yield a lower Hausdorff distance.
  
  % In this work, we choose Hausdorff distance because 
  
  % most global planners in existing geometric navigation systems only plan 2D trajectories, without robot orientation and precise temporal information. 
  % \zichao{better refinemet later}

\subsection{Geometric Navigation Systems}
We benchmark the social compliance of four publicly available geometric navigation systems on \textsc{scand}:

\subsubsection{\texttt{move\_base}~\cite{movebase}}
The Robot Operating System (\textsc{ros}) \texttt{move\_base} is a well-known geometric navigation system that consists of a global planner and a local planner. The global planner uses Dijkstra's algorithm to plan an optimal trajectory on a static costmap and the local planner then employs the Dynamic Window Approach (\textsc{dwa})~\cite{fox1997dynamic} to generate real-time actions given the global trajectory.

% representation of the environment and Dijkstra's algorithm to generate an optimal path from the robot's current pose to the goal. The resulting path is smoothed and interpolated so that the local planner can follow. 
% The \texttt{move\_base} default local planner, the Dynamic Window Approach (\textsc{dwa}) \cite{fox1997dynamic}, operates reactively, considering the robot state, sensor information, kinematic constraints, and global path. It generates real-time linear and angular velocity commands by evaluating various trajectories within the robot's dynamic window, ensuring progress towards the goal while avoiding obstacles.  

\subsubsection{\texttt{move\_base} with social layer (\texttt{move\_base}(s))~\cite{social_layer}}
This method enhances \texttt{move\_base} by incorporating a social layer into the static costmap. It leverages real-time LiDAR scans to identify humans and surrounds them with a Gaussian filter for improved social awareness in planning.

\subsubsection{Human-Aware Planner~\cite{kollmitz15ecmr}} 
The Human-Aware Planner introduces a social cost function into path planning to prioritize human comfort and ensure politeness and safety. It uses time-dependent and kinodynamic path planning to consider the spatial relationship of the robot and humans. 

\subsubsection{CoHAN~\cite{singamaneni2021human}} 
CoHAN is a human-aware navigation planner that uses an extension of the Human-Aware Timed Elastic Band (HATEB) planner~\cite{singamaneni2020hateb} as the local planner to handle complex and crowded navigation scenarios. This system is developed over the \textsc{ros} navigation stack by introducing human safety and human visibility costmap layers into both global and local costmap.

\begin{figure}[t]
    \centering
    \includegraphics[width=0.8\columnwidth]{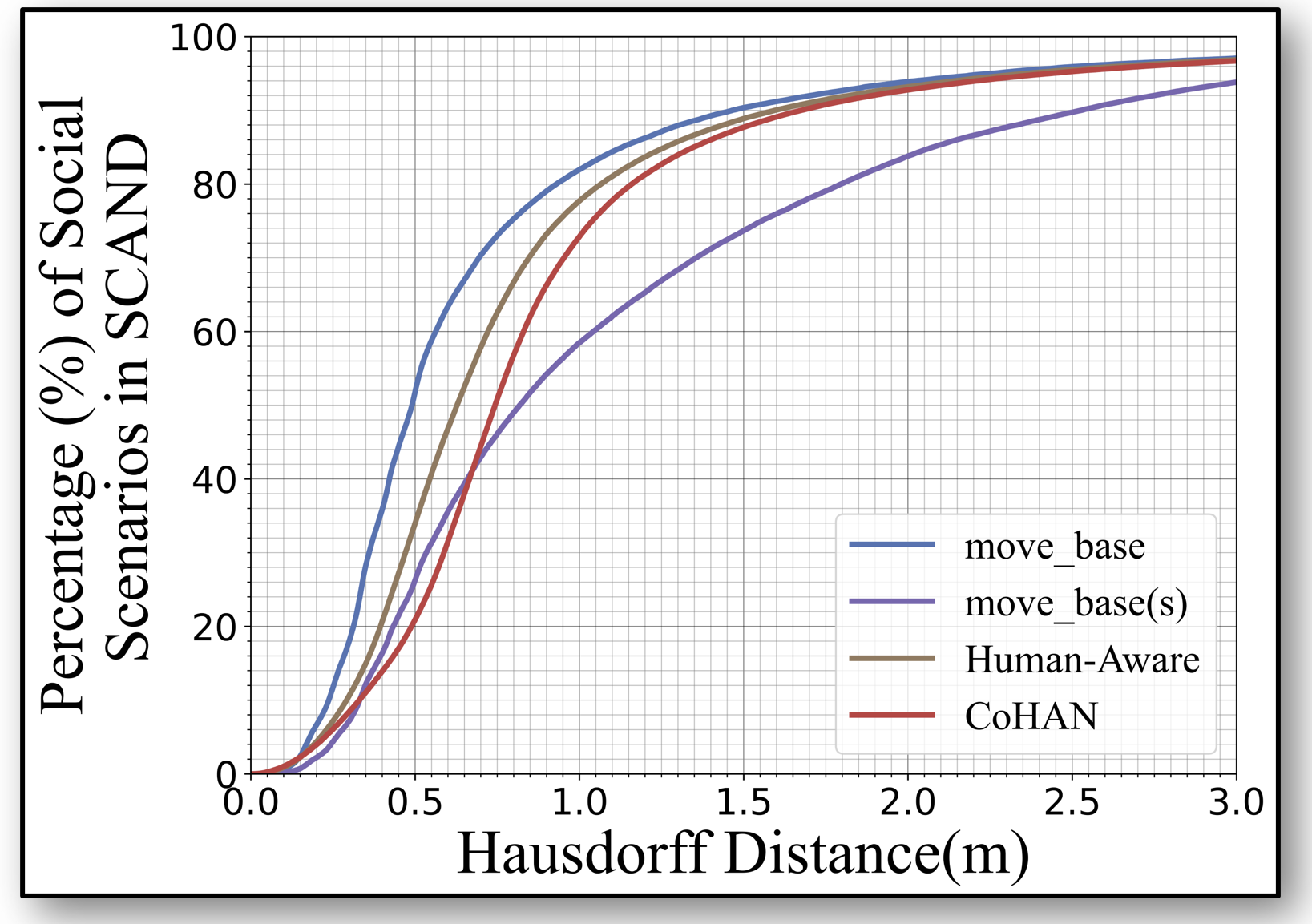}
    \caption{Cumulative Distribution Function (CDF) curves illustrating the Hausdorff distance and social scenario percentages for four distinct geometric planners, analyzed on the \textsc{scand}.}
    \label{fig:ClassicalPlannersComp}
\end{figure}

% \begin{figure*}[t]
%     \centering
%     \includegraphics[width=\linewidth]{figure/figure3.png}    
%     \caption{Case Study Results: Cumulative Distribution Function (CDF) curves of different navigation planners compared against human demonstrations (Hausdorff distance for global planners and L-2 norm for local planners) with in-distribution (\textsc{scand}) and out-of-distribution data.\zichao{1. remove the leftmost figure 2. create a single figure with only movebase, movebase-s, human-aware, cohan, 3. create a two subfigure with movebase+BC+Hybrid (in distribution on the left and out-of-distribution on the right)} }
%     \label{fig:case_study}
% \end{figure*}

We maintain the default parameterizations and configurations for all four geometric navigation systems. Each system is evaluated by playing back \textsc{scand} data, recording the planned trajectories $\mathcal{B}_t$, and comparing them against the \textsc{scand} demonstration trajectories $\mathcal{B}_t^D$ to evaluate their social compliance under Def.~\ref{def: social_compliant}.

Fig.~\ref{fig:ClassicalPlannersComp} illustrates four Cumulative Distribution Function (CDF) curves comparing the performance of the four geometric navigation systems, indicating consistent trends among the systems: across many social scenarios in \textsc{scand}, they manage to maintain a moderate deviation from human demonstrations within $\epsilon = 1.0$. 
We find that \texttt{move\_base} achieves the best alignment with human demonstrations in over $80\%$ of \textsc{scand} navigation scenarios. This outcome is unexpected, given that \texttt{move\_base} is the only navigation system among those evaluated that does not explicitly incorporate social factors into its algorithm. This observation suggests that a general-purpose navigation planner even without tuning for specific social scenarios can often produce socially compliant trajectories.

 % Through an analysis of various social scenarios generated by the navigation system, which include both socially compliant and socially incompliant trajectories, our findings reveal that a significant majority (80\%) of the scenarios intended for social navigation, as represented by \textsc{scand}, are straightforward for a general purpose navigation system, like \texttt{move\_base}, to determine an optimal trajectory. This suggests that the complexity introduced by social navigation requirements can often be effectively managed by existing navigational algorithms.

Additionally, we observe a small portion (roughly 5\%) of scenarios where the geometric planners deviate significantly from human demonstrations (more than 3 meters Hausdorff distance). To understand the causes of such deviations, we sample 50 of these scenarios and inspect the interactions between the robot and the environment. We discover that these scenarios often require sophisticated reasoning skills that a geometric planner does not possess. For example, Fig.~\ref{fig:cover} shows two common social scenarios that are challenging for geometric planners and inevitably lead to poor compliance.  

\subsection{Comparing Geometric and Learning-Based Planner}
\begin{figure}[t]
    \centering
    \includegraphics[width=1\columnwidth]{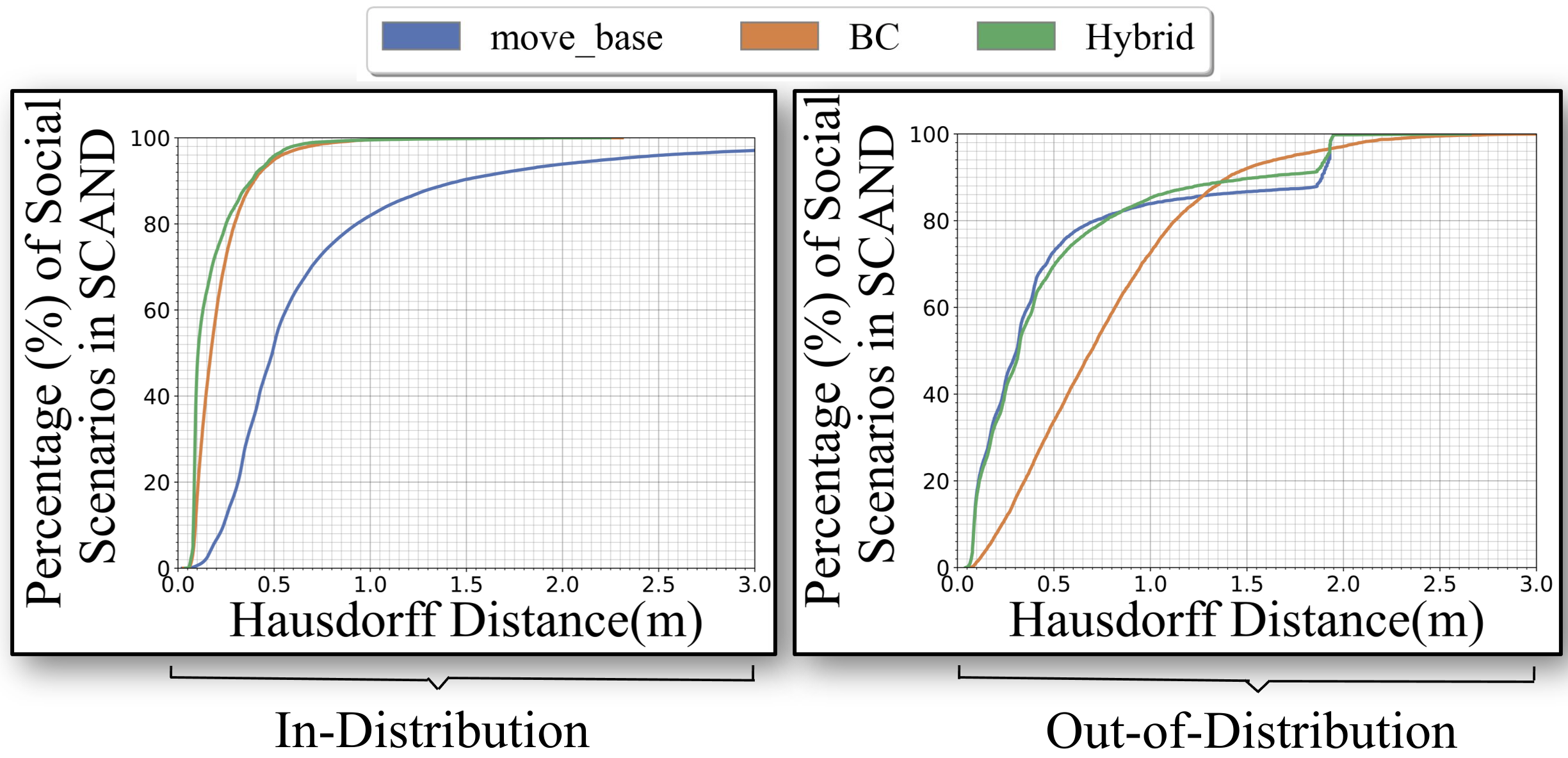}
    \caption{CDF curves of Geometric, Learning-Based, and Proposed Hybrid Navigation Planners on In-Distribution \textsc{scand} and Out-of-Distribution Test Data.}
    \label{fig:InOutDistribution}
\end{figure}
In social robot navigation, human behaviors are unpredictable and the environment can vary significantly at different locations and times. Hence, it is challenging to create one dataset to capture the full distributions of social scenarios. This limitation could cause learning-based methods to suffer from the distribution-shift problem~\cite{domain_shift_Triess2021ASO}. 

In order to investigate this problem, in addition to training and testing on \textsc{scand}, we follow \textsc{scand}'s procedure and collect extra demonstrations as an out-of-distribution test set. These demonstrations contain manually curated social scenarios in a lab environment, including intersection encounter, frontal approach, and people following. We then train a simple learning-based method, Behavior Cloning (BC)~\cite{pomerleau1989alvinn}, on the \textsc{scand} training set and evaluate its performance on both the in-distribution \textsc{scand} test set and the manually curated out-of-distribution test set. Finally, we compare it with a general-purpose geometric planner, \texttt{move\_base}, on both in-distribution and out-of-distribution data. 

We plot the CDFs of the percentage of social scenarios over different Hausdorff distance thresholds in Fig.~\ref{fig:InOutDistribution}. We observe that while BC can outperform \texttt{move\_base} on the in-distribution \textsc{scand} test data, BC's performance significantly deteriorates on the out-of-distribution test set. \texttt{move\_base}'s performance remains unchanged. This result indicates that training on the \textsc{scand} dataset might be affected by the distribution-shift problem, leading to poor generalization when facing various unseen social scenarios.

\section{Rethinking Social Robot Navigation}
\label{sec::approach}

% \subsection{Leveraging the Best of Both Worlds}
% \begin{figure}[t]
%     \centering
%     \includegraphics[width=8.5cm]{figure/1-2F.png}
%     \caption{}
%     \label{fig:hybrid_planner_qualitative}
% \end{figure}

Our case study results suggest 
\begin{enumerate*}
    \item geometric navigation systems can produce socially compliant navigation behaviors in many social scenarios but will inevitably fail in certain social scenarios; and
    \item learning-based methods (in our case, BC) have the potential to solve challenging social scenarios where geometric approaches fail but using them alone can suffer from distribution-shift problems. 
\end{enumerate*}

Our case study motivates us to rethink the approach to solving social robot navigation: can we take advantage of both geometric and learning-based approaches? In this work, we make an initial exploratory effort to answer this question: we develop a hybrid planner that uses a geometric navigation system as the backbone and complements it with a learned model (BC) for handling difficult social navigation scenarios. We also train a classifier to serve as a gating function, determining when \texttt{move\_base} is likely to fail at generating a socially compliant behavior, at which point we switch to using the output from BC. To be specific, let $S_t$ be a social scenario at time step $t$. We instantiate our hybrid navigation planner $\mathcal{F}(\cdot)$ based on a geometric navigation planner $\mathcal{C}(\cdot)$, a learning-based planner $\mathcal{L}_\theta(\cdot)$ with learnable parameters $\theta$, and a gating function $\mathcal{G}_\phi(\cdot)$ with learnable parameters $\phi$ that selects between the output from the geometric and learning-based planners: 
\[
    \mathcal{B}_t = \mathcal{F}(\mathcal{S}_t) = \mathcal{G}_{\phi}(\mathcal{C}(\mathcal{S}_t), \mathcal{L}_\theta(\mathcal{S}_t), \mathcal{S}_t). 
\]
The parameters $\phi$ and $\theta$ can be learned using supervised learning on the navigation scenario and behavior tuples $\{\mathcal{S}_t, \mathcal{B}_t^D\}_{t=1}^T$ in \textsc{scand}: 
\[
\begin{split}
    &\argmin_{\phi, \theta}  \sum_{t=1}^T d(\mathcal{S}_t), \\ 
    &d(\mathcal{S}_t) = \lVert \mathcal{B}_t - \mathcal{B}_t^D \rVert, \\
    &\mathcal{B}_t = \mathcal{G}_{\phi}(\mathcal{C}(\mathcal{S}_t), \mathcal{L}_\theta(\mathcal{S}_t), \mathcal{S}_t).
    \label{eqn::argmin}
\end{split}
\]

Among the many ways to learn $\mathcal{G}_\phi(\cdot)$ and $\mathcal{L}_\theta(\cdot)$ (either jointly or separately), in this work, we present a simple implementation that first learns a classifier $\mathcal{M}_\phi(\mathcal{S}_t)$ based on the difference $d$ between $\mathcal{B}_t^D$ and $\mathcal{C}(\mathcal{S}_t)$ to choose between $\mathcal{C}(\mathcal{S}_t)$ and $\mathcal{L}_\theta(\mathcal{S}_t)$: 
\[
    \mathcal{B}_t = 
\begin{cases}
    \mathcal{C}(\mathcal{S}_t),& \text{if } \mathcal{M}_\phi(\mathcal{S}_t) = 1,\\
    \mathcal{L}_\theta(\mathcal{S}_t),& \text{if } \mathcal{M}_\phi(\mathcal{S}_t) = 0.
\end{cases}
\]
$\mathcal{C}(\cdot)$ can already produce socially compliant behaviors when $d \leq \epsilon$, while $\mathcal{L}_\theta(\cdot)$ only learns to address navigation scenarios where $d > \epsilon$.
By comparing $\mathcal{C}(\mathcal{S}_t)$ against $\mathcal{B}_t^D$, i.e., $d = ||\mathcal{C}(\mathcal{S}_t) - \mathcal{B}_t^D||$, we separate the original \textsc{scand} $\mathcal{D}$ into a socially compliant, $\mathcal{D}^C$, and a socially non-compliant, $\mathcal{D}^N$, subset with respect to $\mathcal{C}(\cdot)$, and form a supervised dataset $\{\mathcal{S}_t, c_t\}_{t=1}^T$, in which $c_t = 1$ if $\mathcal{S}_t \in \mathcal{D}^C$ ($d \leq \epsilon$) and $c_t = 0$ if $\mathcal{S}_t \in \mathcal{D}^N$ ($d > \epsilon$). Then, $\mathcal{M}_\phi(\cdot)$ is learned via supervised learning with a cross-entropy loss to classify whether $\mathcal{C}(\mathcal{S}_t)$ is socially compliant or not:  
\[
\tiny
    \phi^* = \argmax_\phi \sum_{t=1}^T \log \frac{\exp(\mathcal{M}_\phi(\mathcal{S}_t)[c_t])}{\exp{(\mathcal{M}_\phi(\mathcal{S}_t)[0])}+\exp{(\mathcal{M}_\phi(\mathcal{S}_t)[1])}}.
\]
The learning-based planner $\mathcal{L}_\theta$ is then learned to minimize the difference between its outputs and demonstrations in $\mathcal{D}^N$: 
\[
    \theta^* = \argmin_{\theta} \sum_{(\mathcal{S}_t, \mathcal{B}_t^D) \in \mathcal{D}^N} ||\mathcal{L}_\theta(\mathcal{S}_t) - \mathcal{B}_t^D||. 
    \label{eqn:bc}
\]
During deployment, $\mathcal{F}(\cdot)$ first uses $\mathcal{M}_{\phi^*}(\cdot)$ to classify if $\mathcal{C}(\mathcal{S}_t)$ is socially compliant or not, and then executes $\mathcal{C}(\mathcal{S}_t)$ if compliant or $\mathcal{L}_{\theta^*}(\mathcal{S}_t)$ if not. 

We evaluate the hybrid planner on both the in-distribution and out-of-distribution test sets and compare it with the geometric and BC planners. 
% Fig.~\ref{fig:hybrid_planner_qualitative} shows an example of the qualitative result of the planned trajectory. 
Fig.~\ref{fig:InOutDistribution} shows the hybrid planner performs similarly to the best-performing planner on either test set, suggesting that the hybrid planner can take advantage of the best of both worlds during the playback tests.

\section{Physical Experiments}
\label{sec::experiments}

\begin{figure}[t]
    \centering
    \resizebox{0.45\textwidth}{!}{
        \input{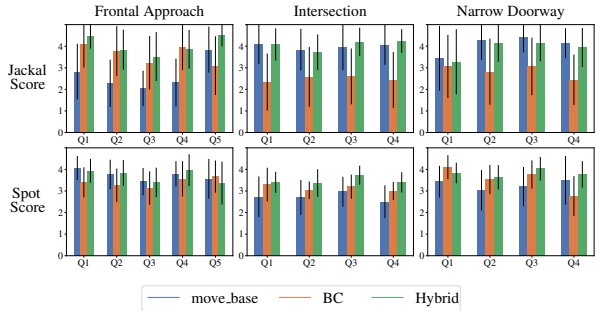}
    }
    \caption{Human Study Average Scores Per Question.}
    \label{fig::physical}
\end{figure}

\begin{table}[t]
    \centering
    \begin{tabular}{cccc}
        \toprule
        Jackal & Frontal & Intersection & Doorway \\ 
        \midrule
        \texttt{move\_base} & $2.66\pm0.64$ & $3.98\pm0.10$ & $\mathbf{4.08\pm0.38}$\\
        
        BC & $3.63\pm0.40$ & $2.49\pm0.11$ & $2.84\pm0.25$\\
        
        Hybrid & $\mathbf{4.04\pm0.39}$ & $\mathbf{4.06\pm0.20}$ & $3.89\pm0.36$\\
        \midrule
        Spot & Frontal & Intersection & Doorway \\ 
        \midrule
        
        \texttt{move\_base} & $\mathbf{3.73\pm0.22}$ & $2.72\pm0.17$ & $3.29\pm0.19$ \\
        
        BC & $3.41\pm0.19$ & $3.13\pm0.12$ & $3.54\pm0.49$ \\

        Hybrid & $3.70\pm0.26$ & $\mathbf{3.48\pm0.15}$ & $\mathbf{3.82\pm0.14}$ \\
        
        \bottomrule
    \end{tabular}
    \caption{Human Study Average Scores Per Method and Scenario: Participants generally prefer the robot with the hybrid approach to the pure classical or the pure BC approach.}
    \label{tab::physical}
\end{table}

We conduct a human study in a series of physical experiments to assess the social compliance of our proposed hybrid approach, in comparison to an existing classical planner, i.e., \texttt{move\_base}, and an end-to-end learning-based method, i.e., BC trained on \textsc{scand}. The experiments are conducted using a wheeled Clearpath Jackal and a legged Boston Dynamics Spot to show the generalizability of our proposed hybrid approach to robots with different morphologies on two university campuses, George Mason University (GMU) and The University of Texas at Austin (UT), respectively. 

We test the robots' social compliance within three distinct social scenarios, i.e., Frontal Approach, Intersection, and Narrow Doorway. 
We keep the same setup of our hybrid approach among all three scenarios. 
The three methods are randomly shuffled and repeated five times, and human participants are requested to respond to a questionnaire with 4-5 questions using Likert scales \cite{pirk2022protocol, xiao2022learning} following each run. Each scenario is tested on ten different individuals (fifteen interactions per individual).  

\subsection{Social Compliance Questionnaire}
For Frontal Approach, the five questions are\footnote{$^*$ denotes negatively formulated questions, for which we reverse-code the ratings to make them comparable to the positively formulated ones.}: 
\begin{enumerate}
    \item \emph{The robot moved to avoid me.}
    \item \emph{The robot obstructed my path$^*$.}
    \item \emph{The robot maintained a safe and comfortable distance at all times.}
    \item \emph{The robot nearly collided with me$^*$.}
    \item \emph{It was clear what the robot wanted to do.}
\end{enumerate}

For Intersection, the four questions are: 
\begin{enumerate}
    \item \emph{The robot let me cross the intersection by maintaining a safe and comfortable distance.}
    \item \emph{The robot changed course to let me pass.}
    \item \emph{The robot paid attention to what I was doing.}
    \item \emph{The robot slowed down and stopped to let me pass.}
\end{enumerate}

For Narrow Doorway, the four questions are: 
\begin{enumerate}
    \item \emph{The robot got in my way$^*$.}
    \item \emph{The robot moved to avoid me.}
    \item \emph{The robot made room for me to enter or exit.}
    \item  \emph{It was clear what the robot wanted to do.}
\end{enumerate}
The quantitative results of our experiments are shown in Fig.~\ref{fig::physical}, where we plot the per-question average along with error bars for the three methods in each of the scenarios.

\subsection{Engineering Endeavors During Deployment}
When deploying the hybrid planner on the robots, we incorporate a voting system that employs hysteresis by considering the past $n$ time steps with a threshold value of $r$ to switch between planners. To prevent collisions, we set a strict condition: if the robots get within $p$ meters of an obstacle, they immediately switch to the geometric planner for the next $t$ seconds. We empirically set the values for $n$, $r$, $p$, and $t$ depending on the location and robot. These adjustments make our hybrid system more stable and safer when running on real robots.

\subsection{Jackal Experiments at GMU}
\begin{figure}
    \centering
    \includegraphics[width=\columnwidth]
    {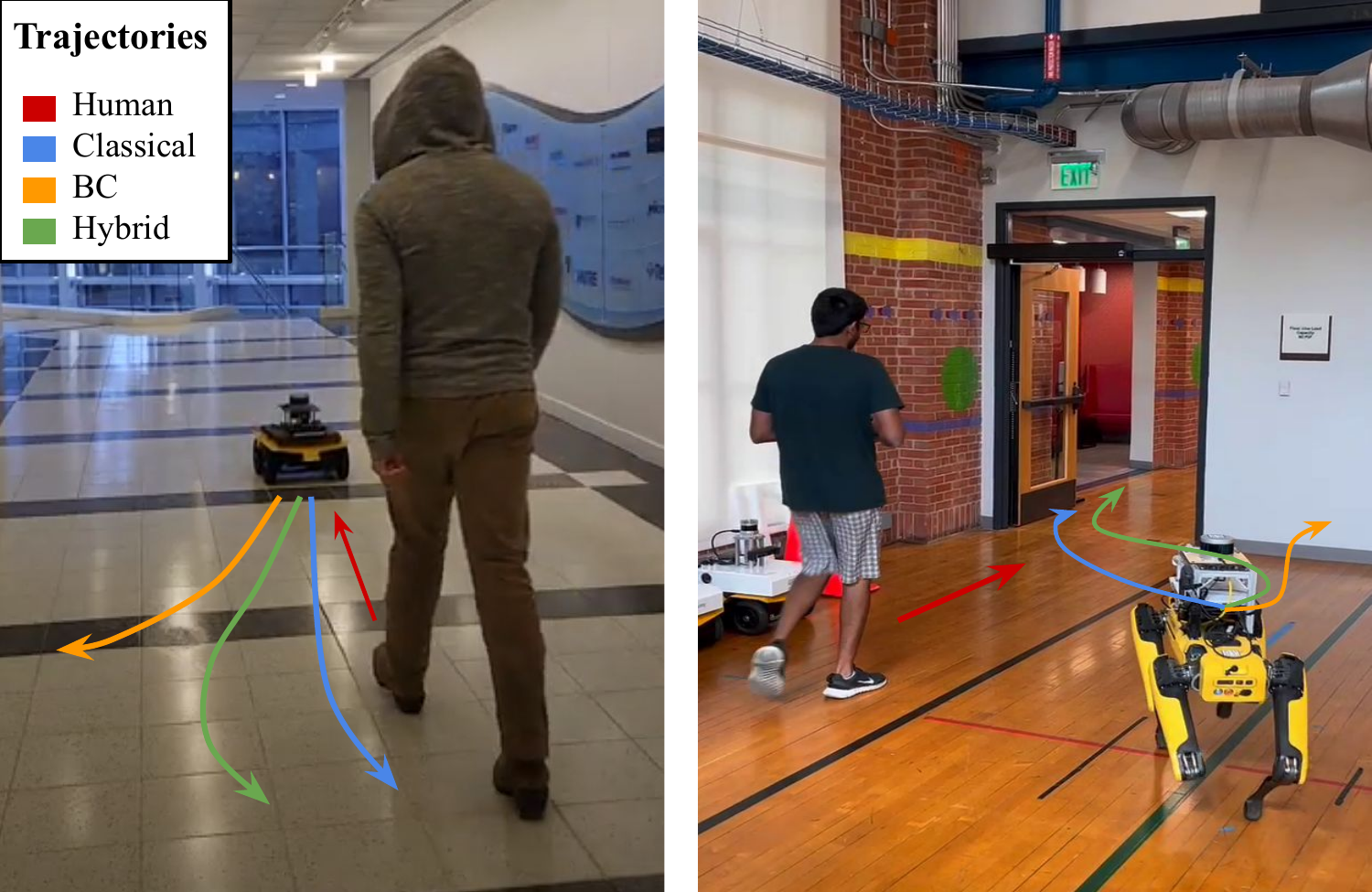}
    \caption{GMU Jackal Frontal Approach (Left) and UT Spot Narrow Doorway (Right) Robot Experiments.}
    \label{fig::robot_exp}
    
\end{figure}

For the experiments at GMU with the wheeled Jackal, our hybrid method shows the most distinguishable social behavior for the frontal approach, maintaining a safe distance while passing by another person coming straight from the opposite side. For the other two scenarios, the classifier of our hybrid method mostly commands to stick to the classical planner and therefore we see similar performance of the classical and our hybrid approach. On the other hand, BC is inconsistent in the majority of the runs: Either it cannot reach the goal successfully, which is evident in the low BC scores for Jackal Frontal Approach Q5 (Fig.~\ref{fig::physical} top left) and Jackal Narrow Doorway Q4 (Fig.~\ref{fig::physical} top right), or we have to manually intervene to avoid an imminent collision with the human subject or surroundings. 
% For negatively formulated questions, we reverse-code the ratings to make them comparable to the positively formulated ones. 
Across the different scenarios, we observe that our approach remains consistent by maintaining the highest average in most of the questions. With both GMU and UT experiments, we ran a one-way ANOVA test on the data from each question with three groups, and the test confirms the statistical significance of the comparison at a 95\% confidence level.
We show the Jackal Frontal Approach experiment in Fig.~\ref{fig::robot_exp} left as an example: Classical approach follows a trajectory which passes very close to the human; BC avoids the human but it cannot recover back to the correct trajectory and gets too close to the wall before we manually intervene; Hybrid approach reacts early by maintaining a safe distance to the human (using BC) and successfully reaches the goal (using \texttt{move\_base}). 

\subsection{Spot Experiments at UT Austin}
We also conduct experiments with a legged Spot at UT. In general, we observe our hybrid approach still performs the most consistently across all three scenarios. However, the classical planner's performance is slightly worse compared to the GMU Jackal experiments. It does not perform well on Spot Intersection (Fig.~\ref{fig::physical} lower middle) and Spot Narrow Doorway (Fig.~\ref{fig::physical} lower right, except Q4 since unlike BC, the classical approach can always reach the goal). We posit this is caused by the different motion morphology of the Spot: legged robots are holonomic, and like humans it is possible for them to side-step during a social interaction. Not being able to do so due to the limitation of \texttt{move\_base} may cause its movement to be perceived unnatural. BC performs slightly better for Spot in Intersection (Fig.~\ref{fig::physical} lower middle) and Narrow Doorway (Fig.~\ref{fig::physical} lower right). 
We show the Spot Narrow Doorway experiment in Fig.~\ref{fig::robot_exp} right as an example: Classical approach first follows the shortest path until it gets close to the human and avoids the human; BC avoids the human but gets lost thereafter; Hybrid approach can slow down and avoid the human in the beginning and successfully pass the narrow doorway in the end. 
\section{Conclusions, Limitations, and Future Work}
\label{sec::conclusions}
% \zichao{we should mention takeaway as follows
% \begin{enumerate}
%     \item a majority of scenarios geometric navigation system is good
%     \item it is true that geometric planner cannot solve social navigation problem completely due to lack of semantic understanding. But these challenging scenarios are somewhat rare in real-life (as shown by scand)
%     \item although end-to-end learning has shown promises in many different field, due to a lack of data, it is not robust and unpredictable which is bad for real world deployment
%     \item a hybrid approach to use geometric planner majority of the times and invoke learning-based approach is promising and worth more investigation
% \end{enumerate}
% }
This work presents a case study on benchmarking social compliance of different existing geometric navigation systems using the \textsc{scand} dataset. Our case study results suggest 
\begin{enumerate*}
    \item geometric navigation systems can produce socially compliant navigation behaviors in many social scenarios (up to 80\%) but will inevitably fail in certain social scenarios; and
    \item learning-based methods (in our case, BC) have the potential to solve challenging social scenarios where geometric approaches fail but using them alone can suffer from distribution-shift problems. 
\end{enumerate*}
These findings motivate us to rethink social robot navigation: we propose a hybrid paradigm that leverages both geometric-based and learning-based planners. We develop a hybrid planner and evaluate its performance both in the playback tests and on two physical robots as a proof-of-concept experiment. The experiment shows promising results of the hybrid paradigm. 

This work has several limitations. First, getting the system working robustly during deployment poses engineering challenges. It requires further investigation to understand how to integrate geometric-based and learning-based methods better. In addition, the BC algorithm used in this work is a straightforward representative of a variety of learning-based methods, which can be replaced by more sophisticated model designs. Overall, this work serves as a preliminary effort to rethink social robot navigation with a hybrid paradigm and to inspire further investigation of this paradigm in the future.

\section{Acknowledgements}
This work has taken place in the RobotiXX Laboratory at George Mason University, Autonomous Mobile Robotics Laboratory (AMRL), and Learning Agents Research Group (LARG) at The University of Texas at Austin. RobotiXX research is supported by ARO (W911NF2220242, W911NF2320004, W911NF2420027), AFCENT, Google DeepMind, Clearpath Robotics, and Raytheon Technologies. 
AMRL research is supported by NSF (CAREER2046955, IIS-1954778, SHF-2006404), ARO (W911NF-19-2- 0333, W911NF-21-20217), DARPA (HR001120C0031), Amazon, JP Morgan, and Northrop Grumman Mission Systems. 
LARG research is supported by NSF (CPS-1739964, IIS-1724157, NRI-1925082), ONR (N00014-18-2243), FLI (RFP2-000), ARO
(W911NF19-2-0333), DARPA, Lockheed Martin, GM, and Bosch. Any opinions, findings, and conclusions expressed in this material are those of the authors and do not necessarily reflect the views of the sponsors.

\bibliographystyle{IEEEtran}
% \footnotesize{
\bibliography{references}
% }
\end{document}